\documentclass[fleqn,12pt]{article}
\usepackage{xspace}
\def\la{($\lambda$)\xspace}
\title{A History of Meta-gradient: \\Gradient Methods for Meta-learning}
\author{Richard S. Sutton}
\date{February 19, 2022}

\begin{document}
\bibliographystyle{apalike}

\maketitle
 
Systems for learning parameters often have meta-parameters such as step sizes, initial weights, or dimensional weightings. With a given setting of the meta-parameters, the learning system is complete and capable of finding parameters that are suited to the task, but its efficiency typically depends on the particular choice of meta-parameters. This has led to interest in learning processes that can find good choices for meta-parameters automatically from experience. These higher-level learning methods are often characterized as ``learning to learn'' or, as we shall call them here, \textit{meta-learning}. Meta-learning has been explored extensively within machine learning for many years (e.g., see Thrun \& Pratt 1998).
 
A class of meta-learning methods that appears particularly powerful and that has attracted considerable recent attention are those based on stochastic gradient descent or ascent (e.g., Andrychowicz et al.~2016; Finn, Abeel \& Levine 2017; Xu, van Hasselt \& Silver 2018). The rise of deep learning has shown that gradient methods can be surprisingly effective in the base learning system (e.g., AlphaZero (Silver et al.\ 2018), DeepStack (Morav\v c\' ik et al.\ 2017), Alexnet (Krizhevsky et al.\ 2012)), so it is natural to consider gradient methods for learning meta-parameters. Let us call such gradient methods for meta-learning \textit{meta-gradient methods}, after Xu et al.\ (2018). Almost all meta-gradient methods use a gradient method in the base learning system as well. Meta-gradient methods have shown promise; they are very general and, in some cases, have achieved learning performance equal to that for hand-tuned meta-parameters. Let us briefly consider the origins of meta-gradient learning methods and the major steps leading to recent developments.

The meta-parameter learned in all of the earliest meta-gradient methods was the step size or ``learning rate'' of supervised learning systems. To our knowledge the first meta-gradient method was a method for setting the step-size (gain) meta-parameter of a servomechanism (Sutton 1981). This algorithm adapted only one meta-parameter and, in this sense, did not realize the full potential of automated meta-learning methods. That is, when a person tunes the meta-parameters, they must be few, but if they can be chosen automatically, then there can be many. Jacobs (1988) extended Sutton's work to many step-size parameters, one for each weight of a multi-layer neural network. Schraudolph and Sejnowski (1996) applied Jacobs' algorithm, called \textit{Delta-Bar-Delta} along with momentum and other techniques to accelerate learning in deep networks. A limitation of Delta-Bar-Delta is that it can be applied only to batch training, not incrementally to individual training examples, as in \textit{stochastic} gradient descent. Sutton (1992a; 1992b) generalized Delta-Bar-Delta to an incremental algorithm, called \textit{Incremental Delta-Bar-Delta}, or \textit{IDBD}, and related algorithms, but all of which were limited to linear (one layer) supervised learning. The ideas underlying IDBD were originally and contemporaneously developed as models of biological meta-learning systems (Sutton 1982; Gluck, Glauthier \& Sutton 1992; Schweighofer \& Arbib 1998). Schraudolph (1998, 1999, 2002) extended IDBD to full multi-layer neural networks and refined it in several other ways. Schraudolph and colleagues applied his method, called \textit{Stochastic Meta Descent}, or \textit{SMD}, to good effect in many problem areas, including independent component analysis (Schraudolph \& Giannakopoulos 2000), visual hand and body tracking (Bray et al.\ 2005, 2007a, 2007b; Kehl \& Van Gool 2006), conditional random fields (Vishwanathan et al.\ 2006a), and support vector machines in reproducing kernel Hilbert spaces (Karatzoglou et al.\ 2005; Vishwanathan et al.\ 2006b; G\"{u}nter et al.\ 2007; see also He 2009). Others independently applied SMD successfully in modeling turbulent flow (Milano \& Koumoutsakos 2002), in brain computer interfaces (Buttfield, Ferrez \& Mill\'an 2006; Mill\'an et al.\ 2007), in learning in recurrent neural networks better than real-time recurrent learning (Liu \& Elhanany 2007, 2008; Liu 2007), and in natural language processing (Arun et al.\ 2009). A limitation of SMD and IDBD is that they have some parameters of their own (meta-meta-parameters); Mahmood (2010, Mahmood et al.\ 2012) devised a more robust version of IDBD, called \textit{Autostep}, that removed the need to tune any parameters manually. Koop (2008) developed a form of IDBD for logistic rather than linear functions such that it is suited to classification rather than regression.

All of the above meta-gradient methods treated \textit{step-size} meta-parameters, but it was always clear that the ideas were more general. For example, Schaal and Atkeson (1998) readily extended the ideas of IDBD and its derivation techniques beyond step sizes to learn the dimension weightings in learning methods for locally-weighted regression. More recently, Finn, Abbeel, and Levine (2017) introduced a meta-gradient method for learning the starting value of the base parameter each time a new task is encountered (in effect a meta-parameter); they called their method \textit{model-agnostic meta-learning}, or MAML, because it was independent of the form of the base supervised learner. Veeriah, Zhang, and Sutton (2017; Zhang 2018) introduced a meta-gradient method, inspired by IDBD and SMD, that treated the first-layer weights of a two-layer neural network as meta-parameters. Their method, called \textit{crossprop}, drives the hidden units to form a feature representation that generalizes well, as assessed by incremental cross-validation, as opposed to backprop which just tries to reduce the error on the examples seen so far.

Meta-gradient methods have also been used in reinforcement learning. The first may have been that by Yu, Aberdeen, and Schraudolph (2006), who used SMD with policy-gradient reinforcement learning methods. Silver et al.~(2013) apparently used IDBD/SMD in combination with Monte-Carlo, contextual-bandit, and temporal-difference reinforcement-learning methods, but did not provide details or equations for their method. Dabney (2014; Dabney \& Barto, 2012) provided an extensive review of step-size adaptation methods in reinforcement learning but identified no meta-learning methods based on gradient descent.
Nagabandi et al.\ (2018) developed the extension of MAML to reinforcement learning. Particularly ambitious is recent work by Veeriah et al.~(2019) that attempts to use meta-gradients to specify the ``question parameters'' of general value functions used as auxiliary tasks within deep reinforcement learning. 

Considering specifically the use of meta-gradient methods for temporal-difference (TD) learning, the first work may have been that by Bagheri, Thill, Koch, and Konen (2015), who combined IDBD with one-step TD learning. Thill (2015) then extended that work to eligibility traces (TD\la) and to nonlinear transfer functions (as in Koop 2008) to produce an algorithm he called \textit{nl-IDBD\la}. There are several choices in the extension to eligibility traces. Kearney et al.~(2018) made different choices to derive a slightly different extension of IDBD to TD\la with they called \textit{TIDBD}. TIDBD was applied to a variety of robotic prediction and control problems by G\"{u}nther and others (2019a, 2019b). Javed (personal communication) argues that Thill's extension to eligibility traces is more effective than that proposed by Kearney et al.
Young et al.\ (2018) developed yet a third way of combining IDBD and TD\la (and general nonlinearity) to produce their algorithm, called \textit{Metatrace}.
 

Meta-gradient learning shows promise in tuning the learning parameters of reinforcement learning methods; as of yet, the area remains relatively unexplored. The strengths and sensitivities of these methods have yet to be carefully examined and there is much work to be done in developing meta-gradient methods suited for reinforcement learning problems. 

\section*{References}

\parskip=5pt
\parindent=0pt
\def\hangin{\hangindent=0.2in}
\def\bibitem[#1]#2{\hangin}

\hangin
Andrychowicz, M., Denil, M., Gomez, S., Hoffman, M. W., Pfau, D., Schaul, T., Shillingford, B., De Freitas, N. (2016). Learning to learn by gradient descent by gradient descent. In \emph{Advances in Neural Information Processing Systems} (pp.~3981--3989).

\hangin
Arun, A., Dyer, C., Haddow, B., Blunsom, P., Lopez, A., Koehn, P. (2009). Monte Carlo inference and maximization for phrase-based translation. In \emph{Proceedings of the Thirteenth Conference on Computational Natural Language Learning} (pp.~102--110). Association for Computational Linguistics.
 
\hangin
Bagheri, S., Thill, M., Koch, P., Konen, W. (2015). Online adaptable learning rates for the game Connect-4. \emph{IEEE Transactions on Computational Intelligence and AI in Games}.
 
\hangin
Bray, M., Koller-Meier, E., M\"{u}ller, P., Schraudolph, N. N., Van Gool, L. (2005). Stochastic optimization for high-dimensional tracking in dense range maps. \textit{IEE Proceedings---Vision, Image and Signal Processing, 152}(4), 501--512.

\hangin
Bray, M., Koller-Meier, E., Schraudolph, N. N., Van Gool, L. (2007a). Fast stochastic optimization for articulated structure tracking. \textit{Image and Vision Computing, 25}(3), 352--364.

\hangin
Bray, M., Koller-Meier, E., Van Gool, L. (2007b). Smart particle filtering for high-dimensional tracking. \textit{Computer Vision and Image Understanding, 106}(1), 116--129.

\hangin
Buttfield, A., Ferrez, P. W., Millan, J. R. (2006). Towards a robust BCI: Error potentials and online learning. \emph{IEEE Transactions on Neural Systems and Rehabilitation Engineering, 14}(2), 164--168.
 
\hangin
Dabney, W. C. (2014). \emph{Adaptive Step-sizes for Reinforcement Learning}. University of Massachusetts at Amherst PhD thesis.

\hangin
Dabney, W., Barto, A. G. (2012). Adaptive step-size for online temporal difference learning. In \emph{Twenty-Sixth Conference of the Association for the Advancement of Artificial Intelligence}.

\hangin
Finn, C., Abbeel, P., Levine, S. (2017). Model-agnostic meta-learning for fast adaptation of deep networks. In \emph{Proceedings of the 34th International Conference on Machine Learning} (pp.~1126--1135).

\hangin
Gluck, M., Glauthier, P., Sutton, R.S. (1992). Adaptation of cue-specific learning rates in network models of human category learning, \emph{Proceedings of the Fourteenth Annual Conference of the Cognitive Science Society}, pp.~540--545, Erlbaum.

\hangin
G\"{u}nter, S., Schraudolph, N. N., Vishwanathan, S. V. N. (2007). Fast iterative kernel principal component analysis. \emph{Journal of Machine Learning Research, 8}, 1893--1918.

\hangin
G\"{u}nther, J., Ady, N. M., Kearney, A., Dawson, M. R., Pilarski, P. M. (2019a). Examining the Use of Temporal-Difference Incremental Delta-Bar-Delta for Real-World Predictive Knowledge Architectures. ArXiv:1908.05751.

\hangin
G\"{u}nther, J., Kearney, A., Ady, N. M., Dawson, M. R., Pilarski, P. M. (2019b). Meta-learning for predictive knowledge architectures: a case study using TIDBD on a sensor-rich robotic arm. In \emph{Proceedings of the 18th International Conference on Autonomous Agents and MultiAgent Systems} (pp.~1967--1969). International Foundation for Autonomous Agents and Multiagent Systems.

\hangin
He, W. (2009). Limited stochastic meta-descent for kernel-based online learning. \emph{Neural Computation, 21}(9), 2667--2686.
 
\hangin
Jacobs, R. A. (1988). Increased rates of convergence through learning rate adaptation. \emph{Neural Networks, 1}(4):295--307.

\hangin
Karatzoglou, A., Vishwanathan, S. V. N., Schraudolph, N. N., Smola, A. J. (2005). Step size-adapted online support vector learning. In \emph{ISSPA} (pp.~823--826).

\hangin
Kearney, A., Veeriah, V., Travnik, J. B., Sutton, R. S., Pilarski, P. M. (2018). TIDBD: Adapting Temporal-difference Step-sizes Through Stochastic Meta-descent. ArXiv: 1804.03334.

\hangin
Kehl, R., Van Gool, L. (2006). Markerless tracking of complex human motions from multiple views. \textit{Computer Vision and Image Understanding, 104}(2--3), 190--209.
 
\hangin
Koop, A. (2008). Investigating experience: Temporal Coherence and Empirical Knowledge Representation. University of Alberta MSc thesis.
 
\hangin
Krizhevsky, A., Sutskever, I., Hinton, G. E. (2012). Imagenet classification with deep convolutional neural networks. In: \emph{Advances in Neural Information Processing Systems, 25}.

\hangin
Liu, Z. (2007). Hardware-Efficient Scalable Reinforcement Learning Systems. University of Tennessee PhD thesis.
 
\hangin
Liu, Z., Elhanany, I. (2007). Fast and scalable recurrent neural network learning based on stochastic meta-descent. In \emph{2007 American Control Conference} (pp.~5694--5699). IEEE.
 
\hangin
Liu, Z., Elhanany, I. (2008). A fast and scalable recurrent neural network based on stochastic meta descent. \emph{IEEE Transactions on Neural Networks, 19}(9), 1652--1658.
 
\hangin
Mahmood, A.R. (2010). \emph{Automatic Step-size Adaptation In Incremental Supervised Learning}. University of Alberta MSc thesis.

\hangin
Mahmood, A. R., Sutton, R. S., Degris, T.,  Pilarski, P. M. (2012). Tuning-free step-size adaptation. In \emph{Proceeedings of the 2012 IEEE International Conference on Acoustics, Speech and Signal Processing (ICASSP)}, pp.~2121--2124. IEEE.

\hangin
Milano, M., Koumoutsakos, P. (2002). Neural network modeling for near wall turbulent flow. \emph{Journal of Computational Physics, 182}(1), 1--26.
 
\hangin
Mill\' an, J. D. R., Buttfield, A., Vidaurre, C., Cabeza, R., Krauledat, M., 
Schl\"{o}gl, A., Shenoy, P., Blankertz, B., Rao, R. P. N., Cabeza, R., Pfurtscheller, G., M\"{u}ller, K.-R. (2007). Adaptation in brain-computer interfaces. In \emph{Toward Brain-Computer Interfacing}.
 
\hangin
Morav\v c\'ik, M., Schmid, M., Burch, N., Lis\'y, V., Morrill, D., Bard, N., ..., Bowling, M. (2017). Deepstack: Expert-level artificial intelligence in heads-up no-limit poker. \emph{Science, 356}(6337), 508--513.

\hangin
Nagabandi, A., Clavera, I., Liu, S., Fearing, R. S., Abbeel, P., Levine, S., Finn, C. (2018). Learning to adapt in dynamic, real-world environments through meta-reinforcement learning. ArXiv:1803.11347.

\hangin
Schaal, S., Atkeson, C. G. (1998). Constructive incremental learning from only local information. \emph{Neural Computation, 10}(8), 2047--2084.

\hangin
Schraudolph, N. (1998). Online local gain adaptation for multi-layer perceptions. Technical report/IDSIA-09-98.

\hangin
Schraudolph, N. N. (1999). Local gain adaptation in stochastic gradient descent. In \emph{Proceedings of the International Conference on Artificial Neural Networks}, pp.~569--574. IEEE, London. Also Technical Report IDSIA-09-99.

\hangin
Schraudolph, N. N. (2002). Fast curvature matrix-vector products for second-order gradient descent. \emph{Neural Computation, 14}(7):1723--1738.

\hangin
Schraudolph, N. N., Giannakopoulos, X. (2000). Online independent component analysis with local learning rate adaptation. In \textit{Advances in Neural Information Processing Systems} (pp.~789--795).

\hangin
Schraudolph, N. N., Sejnowski, T. J. (1996). Tempering backpropagation networks: Not all weights are created equal. In \emph{Advances in Neural Information Processing Systems} (pp.~563--569).

\hangin
Schraudolph, N. N., Yu, J., Aberdeen, D. (2006). Fast online policy gradient learning with SMD gain vector adaptation. In \emph{Advances in Neural Information Processing Systems}, pp.~1185--1192.

\hangin
Schweighofer, N., Arbib, M. A. (1998). A model of cerebellar metaplasticity. \emph{Learning and Memory, 4}(5), 421--428.
 
\hangin
Silver, D., Hubert, T., Schrittwieser, J., Antonoglou, I., Lai, M., Guez, A., ..., Hassabis, D. (2018). A general reinforcement learning algorithm that masters chess, shogi, and Go through self-play. \emph{Science, 362}(6419), 1140--1144.

\hangin
Silver, D., Newnham, L., Barker, D., Weller, S., McFall, J. (2013). Concurrent reinforcement learning from customer interactions. In \emph{International Conference on Machine Learning} (pp.~924--932).
 
\hangin
Sutton, R.S. (1981). Adaptation of learning rate parameters.  In: \emph{Goal Seeking Components for Adaptive Intelligence: An Initial Assessment}, by A. G. Barto and R. S. Sutton. Air Force Wright Aeronautical Laboratories Technical Report AFWAL-TR-81-1070. Wright-Patterson Air Force Base, Ohio 45433.

\hangin
Sutton, R.S. (1982). A theory of salience change dependent on the relationship between discrepancies on successive trials on which the stimulus is present. Unpublished report.

\hangin
Sutton, R.S. (1992a). Adapting bias by gradient descent: An incremental version of delta-bar-delta. \emph{Proceedings of the Tenth National Conference on Artificial Intelligence}, pp.~171--176, MIT Press.

\hangin
Sutton, R.S. (1992b). Gain adaptation beats least squares? \emph{Proceedings of the Seventh Yale Workshop on Adaptive and Learning Systems}, pp.~161--166, Yale University, New Haven, CT.
 
\hangin
Thill, M. (2015). Temporal difference learning methods with automatic step-size adaptation for strategic board games: Connect-4 and Dots-and-Boxes. Cologne University of Applied Sciences Masters thesis. 
 

\hangin
Thrun, S., Pratt, L. (1998). \emph{Learning to Learn}. Springer, Boston, MA.

\hangin
Veeriah, V., Hessel, M., Xu, Z., Rajendran, J., Lewis, R. L., Oh, J., van Hasselt, H., Silver, D., Singh, S. (2019). Discovery of useful questions as auxiliary tasks. In \emph{Advances in Neural Information Processing Systems} (pp.~9306--9317).

\hangin
Veeriah, V., Zhang, S., Sutton, R. S. (2017). Crossprop: Learning representations by stochastic meta-gradient descent in neural networks. In \emph{Joint European Conference on Machine Learning and Knowledge Discovery in Databases} (pp.~445--459). Springer.

\hangin
Vishwanathan, S. V. N., Schraudolph, N. N., Schmidt, M. W., Murphy, K. P. (2006a). Accelerated training of conditional random fields with stochastic gradient methods. In \emph{Proceedings of the International Conference on Machine Learning} (pp.~969--976).

\hangin
Vishwanathan, S. V. N., Schraudolph, N. N., Smola, A. J. (2006b). Step size adaptation in reproducing kernel Hilbert space. \emph{Journal of Machine Learning Research, 7}, 1107--1133.

\hangin
Xu, Z., van Hasselt, H. P., Silver, D. (2018). Meta-gradient reinforcement learning. In \emph{Advances in Neural Information Processing Systems} (pp.~2396--2407).

\hangin
Young, K., Wang, B., Taylor, M. E. (2018). Metatrace: Online step-size tuning by meta-gradient descent for reinforcement learning control. ArXiv:1805.04514.
 
\hangin
Yu, J., Aberdeen, D., Schraudolph, N. N. (2006). Fast online policy gradient learning with SMD gain vector adaptation. In \emph{Advances in Neural Information Processing Systems} (pp.~1185--1192).
 
\hangin
Zhang, S. (2018). \emph{Learning with Artificial Neural Networks}. University of Alberta MSc thesis.
 
\end{document}